\documentclass[10pt,twocolumn,letterpaper]{article}

\usepackage{cvpr}              

\usepackage{multirow}
\usepackage{colortbl}
\definecolor{bestcell}{HTML}{AED6F1}
\definecolor{sbestcell}{HTML}{D6EAF8}
\newcommand{\best}[1]{\cellcolor{bestcell}\textbf{#1}}
\newcommand{\sbest}[1]{\cellcolor{sbestcell}#1}
\definecolor{robustcell}{HTML}{ABEBC6}
\newcommand{\rbest}[1]{\cellcolor{robustcell}\textbf{#1}}

\renewcommand{\paragraph}[1]{\vspace{.5em}\noindent\textbf{#1}}

\definecolor{cvprblue}{rgb}{0.21,0.49,0.74}
\usepackage[pagebackref,breaklinks,colorlinks,allcolors=cvprblue]{hyperref}

\def\paperID{*****} 
\def\confName{CVPR}
\def\confYear{2026}

\title{DINOv3 Beats Specialized Detectors: A Simple Foundation Model Baseline for Image Forensics}

\author{%
Jieming Yu$^{1}$ \quad
Qiuxiao Feng$^{1}$ \quad
Zhuohan Wang$^{2}$ \quad
Xiaochen Ma$^{1}$\thanks{Corresponding author: \texttt{xiaochen.ma.cs@gmail.com}}\\
$^{1}$The Hong Kong University of Science and Technology \quad
$^{2}$Harvard University
}

\begin{document}
\maketitle
\begin{abstract}


With the rapid advancement of deep generative models, realistic fake images have become increasingly accessible, yet existing localization methods rely on complex designs and still struggle to generalize across manipulation types and imaging conditions.
We present a simple but strong baseline based on DINOv3 with LoRA adaptation and a lightweight convolutional decoder.
Under the CAT-Net protocol, our best model improves average pixel-level F1 by 17.0 points over the previous state of the art on four standard benchmarks using only 9.1\,M trainable parameters on top of a frozen ViT-L backbone, and even our smallest variant surpasses all prior specialized methods.
LoRA consistently outperforms full fine-tuning across all backbone scales. Under the data-scarce MVSS-Net protocol, LoRA reaches an average F1 of 0.774 versus 0.530 for the strongest prior method, while full fine-tuning becomes highly unstable, suggesting that pre-trained representations encode forensic information that is better preserved than overwritten.
The baseline also exhibits strong robustness to Gaussian noise, JPEG re-compression, and Gaussian blur.
We hope this work can serve as a reliable baseline for the research community and a practical starting point for future image-forensic applications.
Code is available at \url{https://github.com/Irennnne/DINOv3-IML}.
\end{abstract}

\section{Introduction}
\label{sec:intro}

Image manipulation poses an increasing threat to the reliability of visual media.
With the rapid development of deep generative models and editing tools, creating realistic fake images has become easier than ever, making visual forgery accessible to a much broader population~\cite{verdoliva2020media}.
This trend has driven growing interest in \emph{image manipulation detection and localization} (IMDL), which aims to determine whether an image has been tampered with and to localize manipulated regions at the pixel level.
Despite substantial progress, IMDL remains challenging in real-world settings due to the diversity of manipulation types, image sources, and post-processing conditions~\cite{verdoliva2020media, ma2024imdlbencocomprehensivebenchmarkcodebase}.

A large body of prior work has approached IMDL through increasingly specialized model designs.
Over time, the field has introduced stronger backbones, more elaborate feature fusion schemes, and various task-specific training strategies to improve localization quality.
While these methods have achieved promising results on standard benchmarks, their gains often do not transfer reliably across datasets and evaluation settings.
Recent benchmark~\cite{ma2024imdlbencocomprehensivebenchmarkcodebase, du2025forensichub} studies have further highlighted this issue, showing that reported improvements can be fragile under unified evaluation and that the field still lacks simple, modern, and reliable baselines for fair comparison.
As a result, the field today faces not only a performance problem, but also a baseline problem: there is still a need for a simple, modern, and reliable starting point that can serve as a stronger foundation for future comparison and development.

In this report, we revisit IMDL from a deliberately simple perspective.
We ask whether a modern vision foundation model can already provide a strong baseline for this task.
We choose DINOv3~\cite{simeoni2025dinov3} because the DINO family has shown strong transfer performance across a broad range of visual tasks, from image-level recognition to dense pixel-level prediction~\cite{oquab2024dinov2, simeoni2025dinov3}.
Building on this foundation, we use DINOv3 with LoRA~\cite{hu2022lora} adaptation, a lightweight convolutional decoder, and a simple boundary-aware loss~\cite{ma2023imlvit}.
The design is intentionally minimal.
We avoid complex task-specific architectural engineering and instead focus on a setup that is easy to reproduce, easy to compare against, and strong enough to serve as a practical default baseline.
This question is particularly relevant for IMDL, since manipulation cues are often subtle, local, and only weakly semantic, making the choice of adaptation strategy especially important when repurposing foundation models for forensic localization.
Beyond the model itself, we therefore examine several simple tuning choices, including backbone scale and adaptation strategy, to clarify what matters most for this task.

We evaluate under two standardized protocols from IMDLBenCo~\cite{ma2024imdlbencocomprehensivebenchmarkcodebase}: the multi-source CAT-Net protocol~\cite{kwon2022catnet} (four benchmarks: CASIAv1~\cite{dong2013casia}, Columbia~\cite{hsu2006columbia}, NIST16~\cite{guan2019nist}, Coverage~\cite{wen2016coverage}) and the more constrained MVSS-Net protocol~\cite{Dong_2023} (CASIAv2-only training~\cite{dong2013casia}, five benchmarks including IMD2020).
Across both protocols, we compare multiple DINOv3 backbone scales and both LoRA and full fine-tuning settings under a unified implementation and checkpoint-selection strategy.
This design allows us to assess not only absolute performance, but also how well different adaptation choices transfer across data regimes.

Empirically, this simple baseline proves highly effective.
Under the CAT-Net protocol, our best configuration improves the average pixel-level F1 score by 17.0 points over the previous state of the art while using only 9.1\,M trainable parameters on top of a frozen ViT-L backbone, and even the smallest variant already surpasses all prior specialized methods like TruFor~\cite{guillaro2023truforleveragingallroundclues} or Mesorch~\cite{mesorch2025}.
Under the more data-constrained MVSS-Net protocol, DINOv3 with LoRA continues to outperform existing methods across all five benchmarks, reaching an average F1 of 0.774 compared with 0.530 for TruFor~\cite{guillaro2023truforleveragingallroundclues}, the strongest prior method.
We also observe a consistent pattern across backbone scales: LoRA performs better than full fine-tuning, especially when supervision is limited.
Under the CASIAv2-only setting, full fine-tuning becomes highly unstable: the smaller backbones train unstably and peak early, while even the largest model remains clearly below its LoRA counterpart.
These results suggest that LoRA better balances adaptation and preservation, retaining the pre-trained features critical for generalization when training data is scarce.
They also indicate that scaling training data may be as important as scaling model capacity, pointing to larger and more diverse IMDL datasets as a promising direction for future work.
Additionally, the baseline exhibits strong robustness to common post-processing perturbations including Gaussian noise, JPEG re-compression, and Gaussian blur, maintaining superior performance over prior methods across all conditions.

Beyond the empirical gains, we believe that establishing such a baseline is valuable in its own right.
Our goal is not only to provide a competitive model, but to raise the default starting point for IMDL research with a simple, effective, and reproducible reference that future work can build on.
We hope this report can serve both as a useful benchmark for academic research and as a practical off-the-shelf foundation for real-world image-forensic applications.

\section{Related Work}
\label{sec:related}

\paragraph{Image Manipulation Localization.}
Classical forensic methods detect manipulation through low-level statistical traces such as noise fingerprints~\cite{cozzolino2018noiseprintcnnbasedcameramodel} and compression artifacts~\cite{kwon2022catnet}.
Recent deep learning approaches have introduced increasingly complex architectures, including multi-view multi-scale fusion~\cite{Dong_2023}, progressive spatial-channel correlation~\cite{liu2022psccnet}, trustworthy localization with reliability estimation~\cite{guillaro2023truforleveragingallroundclues}, multi-spectral attention~\cite{nam2025m2sformermultispectralmultiscaleattention}, pure ViT-based localization~\cite{ma2023imlvit}, and hybrid CNN-Transformer orchestration~\cite{mesorch2025}.
The IMDL-BenCo benchmark~\cite{ma2024imdlbencocomprehensivebenchmarkcodebase} provides standardized evaluation and has exposed significant generalization gaps across these methods.
All of these approaches rely on task-specific backbone design; none leverage general-purpose foundation model features.

\paragraph{Self-Supervised Visual Representations.}
The DINO family~\cite{caron2021dino,oquab2024dinov2,simeoni2025dinov3} has produced increasingly powerful self-supervised ViT encoders.
The original DINO~\cite{caron2021dino} showed that self-distillation without labels gives rise to emergent spatial segmentation in ViT attention maps, revealing that self-supervised objectives naturally encode fine-grained, spatially structured representations.
DINOv2~\cite{oquab2024dinov2} scaled this paradigm with a curated large-scale dataset and demonstrated that frozen features transfer directly to dense prediction tasks—including depth estimation and semantic segmentation—without task-specific fine-tuning.
DINOv3~\cite{simeoni2025dinov3} further introduces gram anchoring to prevent dense feature degradation during extended training schedules, yielding representations with even finer spatial precision.
While this model family has substantially advanced dense prediction benchmarks, its application to image forensics has not been studied; yet the rich spatial priors these features encode seem naturally aligned with the task of detecting subtle statistical discontinuities at manipulation boundaries.

\section{Implementation Details}
\label{sec:method}

\begin{figure*}[t]
  \centering
  \includegraphics[width=\linewidth]{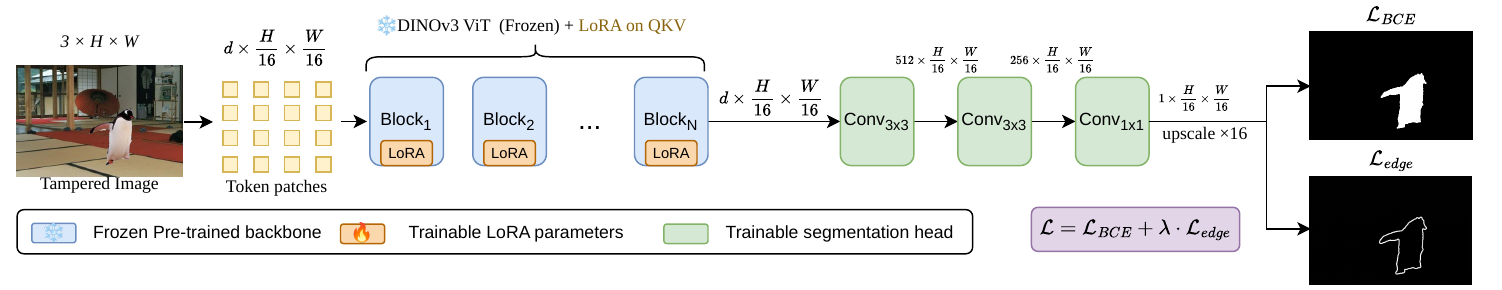}
  \caption{Overview of our framework. A frozen DINOv3 ViT backbone with LoRA injected on QKV projections produces dense patch tokens, which are reshaped into a 2D feature map and decoded by a lightweight convolutional segmentation head into a pixel-level manipulation mask.}
  \label{fig:arch}
\end{figure*}

We investigate the effectiveness of DINOv3~\cite{simeoni2025dinov3} dense features for image manipulation detection and localization (IMDL).
Our framework follows a simple yet principled design (\cref{fig:arch}): a frozen, self-supervised Vision Transformer (ViT) backbone produces spatially dense feature maps, which are then adapted to the forensic task through either low-rank adaptation (LoRA)~\cite{hu2022lora} or full fine-tuning, and decoded into pixel-level manipulation masks by a lightweight convolutional head.
This architecture deliberately keeps the decoder minimal so that any performance differences across configurations can be attributed to the backbone features and adaptation strategy rather than decoder capacity.

\subsection{DINOv3 Feature Extraction}
\label{subsec:backbone}

We evaluate three DINOv3~\cite{simeoni2025dinov3} backbone variants that share a patch size of $16{\times}16$ but differ in model capacity: ViT-S/16 ($d{=}384$), ViT-B/16 ($d{=}768$), and ViT-L/16 ($d{=}1024$), where $d$ denotes the token embedding dimension.
Given an input image of size $H{\times}W$, the backbone produces a spatial feature map $\mathbf{F} \in \mathbb{R}^{\frac{H}{16} \times \frac{W}{16} \times d}$ by extracting the last transformer layer output, reshaping the patch tokens into a 2D grid, and applying layer normalization.
These dense features serve as the sole input to the downstream decoder.

\subsection{Adaptation Strategies}
\label{subsec:adaptation}

We consider two strategies for adapting the pre-trained backbone to the forensic domain.
For \textbf{LoRA adaptation}, we apply LoRA~\cite{hu2022lora} to the QKV projections of every self-attention layer while keeping all original backbone parameters frozen, experimenting with rank configurations $r{=}32$ ($\alpha{=}64$) and $r{=}64$ ($\alpha{=}128$), which yield between 1.4\,M and 12.2\,M end-to-end trainable parameters (LoRA matrices plus the segmentation head) depending on the backbone variant.
For \textbf{full fine-tuning}, we unfreeze all backbone parameters and train them jointly with the decoder, yielding approximately 22\,M, 89\,M, and 309\,M trainable parameters for ViT-S, ViT-B, and ViT-L, respectively.
Comparing the two strategies isolates the effect of parameter-efficient adaptation on forensic generalization.

\subsection{Segmentation Decoder}
\label{subsec:decoder}

The segmentation decoder is a three-layer convolutional head (Conv3) that maps the backbone features to a single-channel manipulation probability map.
Given the feature map $\mathbf{F} \in \mathbb{R}^{\frac{H}{16} \times \frac{W}{16} \times d}$, the decoder applies:
\begin{align}
\mathbf{z}_1 &= \text{ReLU}\!\left(\text{BN}\!\left(\text{Conv}_{3\times3}(\mathbf{F},\; d \!\to\! d/2)\right)\right), \label{eq:dec1} \\
\mathbf{z}_2 &= \text{ReLU}\!\left(\text{BN}\!\left(\text{Conv}_{3\times3}(\mathbf{z}_1,\; d/2 \!\to\! d/4)\right)\right), \label{eq:dec2} \\
\hat{\mathbf{M}} &= \sigma\!\left(\text{Up}\!\left(\text{Conv}_{1\times1}(\mathbf{z}_2,\; d/4 \!\to\! 1)\right)\right), \label{eq:dec3}
\end{align}
where $\text{BN}$ denotes batch normalization, $\sigma$ is the sigmoid function, and $\text{Up}$ denotes bilinear upsampling from $\frac{H}{16}{\times}\frac{W}{16}$ to the original input resolution of $512{\times}512$.
The decoder is intentionally kept lightweight so that detection performance primarily reflects the quality of the backbone features and the chosen adaptation strategy.

\subsection{Loss Function}
\label{subsec:loss}

Training is supervised with a composite loss that combines a global pixel-wise term with an edge-aware term:
\begin{equation}
\mathcal{L} = \mathcal{L}_{\text{BCE}} + \lambda \cdot \mathcal{L}_{\text{edge}},
\label{eq:loss}
\end{equation}
where $\mathcal{L}_{\text{BCE}}$ is the standard binary cross-entropy computed over all pixels, and $\mathcal{L}_{\text{edge}}$ is a weighted binary cross-entropy that up-weights pixels within a boundary region of width 7 pixels around the ground-truth manipulation mask.
We set $\lambda{=}20$ to strongly penalize boundary prediction errors.

\section{Experiments}
\label{sec:experiments}

\subsection{Experimental Setup}
\label{subsec:setup}

\paragraph{Datasets.}
We evaluate on up to five widely adopted IMDL benchmarks that span diverse manipulation types and imaging conditions:
\textbf{CASIAv1}~\cite{dong2013casia} (920 images, splicing and copy-move),
\textbf{Columbia}~\cite{hsu2006columbia} (180 uncompressed spliced images with binary ground-truth masks),
\textbf{NIST16}~\cite{guan2019nist} (564 JPEG-compressed images with complex composite manipulations),
\textbf{Coverage}~\cite{wen2016coverage} (100 copy-move images with visually similar source--target textures), and
\textbf{IMD2020}~\cite{novozamsky2020imd2020} (2{,}010 real-world manipulated images).
All evaluations use the tampered-image-only subset of each dataset.

\paragraph{Training protocols.}
We evaluate under two protocols that differ in training-data scale and diversity, allowing us to assess performance under both large-scale multi-source supervision and constrained single-source training.
The \textbf{CAT-Net protocol}~\cite{kwon2022catnet} uses a balanced mix of CASIA2~\cite{dong2013casia}, FantasticReality~\cite{kniaz2019fantasticreality}, IMD2020~\cite{novozamsky2020imd2020}, and TampCOCO~\cite{kwon2022catnet} for training, and CASIAv1, Columbia, NIST16, and Coverage for testing.
The \textbf{MVSS-Net protocol}~\cite{Dong_2023} is more constrained, using CASIAv2~\cite{dong2013casia} alone (5{,}123 images) for training; all five benchmarks are used for testing, with IMD2020 serving as an out-of-domain set.

\paragraph{Shared training settings.}
All other training hyperparameters are held identical across the two protocols to isolate the effect of the training-data regime.
Images are resized to $512{\times}512$ via bilinear interpolation.
We use the AdamW optimizer with a learning rate of $3 \times 10^{-4}$, a cosine annealing schedule, and 5 warmup epochs over a total of 100 training epochs, with an effective batch size of 240 achieved through gradient accumulation.
Both protocols are implemented within the IMDL-BenCo framework~\cite{ma2024imdlbencocomprehensivebenchmarkcodebase}.

\paragraph{Evaluation metric and checkpoint selection.}
We report pixel-level F1 as the primary metric.
Evaluation is performed every 4 epochs, and for each model we select a single checkpoint whose pixel-level F1, averaged across all test sets of the corresponding protocol, is the highest, avoiding the inflated numbers that arise from per-dataset oracle selection.

\subsection{Results under the CAT-Net Protocol}
\label{subsec:main_results}

\cref{tab:main_results} reports pixel-level F1 scores.
DINOv3-based models substantially outperform all prior methods, with even the smallest variant (ViT-S + LoRA $r{=}32$) surpassing the previous state of the art by 2.7 F1 points and the best configuration (ViT-L + LoRA $r{=}32$) improving by 17.0 points.
Performance scales monotonically with backbone size, and LoRA consistently outperforms full fine-tuning at every scale.

\begin{table*}[t]
\centering
\caption{Pixel-level F1 under the CAT-Net protocol. Prior results are from~\cite{ma2024imdlbencocomprehensivebenchmarkcodebase} and~\cite{mesorch2025}. Best and second-best DINOv3 results per column are highlighted in dark and light blue; the best prior-method result is \textbf{bolded}.}
\label{tab:main_results}
\setlength{\tabcolsep}{7pt}
\begin{tabular}{@{}llccccccc@{}}
\toprule
 & Method & Size & CASIAv1 & Columbia & NIST16 & Coverage & Avg.\ F1 \\
\midrule
\multirow{5}{*}{\rotatebox[origin=c]{90}{\small Prior work}}
 & MVSS-Net~\cite{Dong_2023}         & $512{\times}512$ & 0.583 & 0.740 & 0.336 & 0.486 & 0.536 \\
 & PSCC-Net~\cite{liu2022psccnet}     & $256{\times}256$ & 0.630 & 0.884 & 0.346 & 0.448 & 0.577 \\
 & CAT-Net~\cite{kwon2022catnet}      & $512{\times}512$ & 0.808 & \textbf{0.915} & 0.252 & 0.427 & 0.601 \\
 & TruFor~\cite{guillaro2023truforleveragingallroundclues} & $512{\times}512$ & 0.818 & 0.885 & 0.348 & 0.457 & 0.627 \\
 & Mesorch~\cite{mesorch2025}         & $512{\times}512$ & \textbf{0.840} & 0.890 & \textbf{0.392} & \textbf{0.586} & \textbf{0.677} \\
\midrule
\multirow{3}{*}{\rotatebox[origin=c]{90}{\small ViT-S}}
 & DINOv3 + LoRA $r{=}32$   & $512{\times}512$ & 0.787 & 0.923 & 0.462 & 0.646 & 0.704 \\
 & DINOv3 + LoRA $r{=}64$   & $512{\times}512$ & 0.803 & 0.918 & 0.457 & 0.671 & 0.712 \\
 & DINOv3 + Full Fine-tuning         & $512{\times}512$ & 0.704 & 0.887 & 0.400 & 0.531 & 0.630 \\
\midrule
\multirow{3}{*}{\rotatebox[origin=c]{90}{\small ViT-B}}
 & DINOv3 + LoRA $r{=}32$   & $512{\times}512$ & 0.840 & \sbest{0.938} & 0.565 & 0.715 & 0.764 \\
 & DINOv3 + LoRA $r{=}64$   & $512{\times}512$ & 0.863 & 0.904 & 0.570 & 0.784 & 0.780 \\
 & DINOv3 + Full Fine-tuning         & $512{\times}512$ & 0.815 & 0.918 & 0.518 & 0.652 & 0.726 \\
\midrule
\multirow{3}{*}{\rotatebox[origin=c]{90}{\small ViT-L}}
 & DINOv3 + LoRA $r{=}32$   & $512{\times}512$ & \sbest{0.907} & \best{0.941} & \best{0.636} & \best{0.905} & \best{0.847} \\
 & DINOv3 + LoRA $r{=}64$   & $512{\times}512$ & \best{0.908} & 0.927 & \sbest{0.633} & \sbest{0.882} & \sbest{0.837} \\
 & DINOv3 + Full Fine-tuning         & $512{\times}512$ & 0.882 & \sbest{0.938} & 0.616 & 0.866 & 0.826 \\
\bottomrule
\end{tabular}
\end{table*}

\subsection{Results under the MVSS-Net Protocol}
\label{subsec:mvss_results}

\cref{tab:mvss_results} reports pixel-level F1 scores under the more constrained MVSS-Net protocol.
Despite training on a single source with over an order of magnitude fewer images, The ViT-L LoRA variants outperform all prior methods across all five benchmarks, while smaller LoRA variants still yield substantial gains in average F1.
LoRA ViT-L achieves an average F1 of 0.774, compared to 0.530 for the strongest prior method (TruFor).
In contrast, Full Fine-tuning configurations collapse in this low-data regime: ViT-S and ViT-B train unstably and peak early (average F1 of 0.221 and 0.110), while ViT-L partially recovers (0.681) but remains well below its LoRA counterpart, confirming that preserving pre-trained representations is critical when training data is scarce.

\begin{table*}[t]
\centering
\caption{Pixel-level F1 under the MVSS-Net protocol. Prior results are from~\cite{ma2024imdlbencocomprehensivebenchmarkcodebase}. Highlighting follows the same convention as \cref{tab:main_results}.}
\label{tab:mvss_results}
\setlength{\tabcolsep}{5pt}
\begin{tabular}{@{}llccccccc@{}}
\toprule
 & Method & Size & Coverage & Columbia & NIST16 & CASIAv1 & IMD2020 & Avg.\ F1 \\
\midrule
\multirow{8}{*}{\rotatebox[origin=c]{90}{\small Prior work}}
 & Mantra-Net~\cite{wu2019mantranet}       & $256{\times}256$ & 0.090 & 0.243 & 0.104 & 0.125 & 0.055 & 0.123 \\
 & MVSS-Net~\cite{Dong_2023}               & $512{\times}512$ & 0.259 & 0.386 & 0.246 & 0.534 & 0.279 & 0.341 \\
 & CAT-Net~\cite{kwon2022catnet}            & $512{\times}512$ & 0.296 & 0.584 & 0.269 & 0.581 & 0.273 & 0.401 \\
 & ObjectFormer~\cite{wang2022objectformer} & $224{\times}224$ & 0.294 & 0.336 & 0.173 & 0.429 & 0.173 & 0.281 \\
 & PSCC-Net~\cite{liu2022psccnet}           & $256{\times}256$ & 0.231 & 0.604 & 0.214 & 0.378 & 0.235 & 0.333 \\
 & NCL-IML~\cite{zhou2022ncliml}            & $512{\times}512$ & 0.225 & 0.446 & 0.260 & 0.502 & 0.237 & 0.334 \\
 & TruFor~\cite{guillaro2023truforleveragingallroundclues} & $512{\times}512$ & 0.419 & \textbf{0.865} & \textbf{0.324} & \textbf{0.721} & 0.322 & \textbf{0.530} \\
 & IML-ViT~\cite{ma2023imlvit}             & $1024{\times}1024$ & \textbf{0.435} & 0.780 & 0.331 & \textbf{0.721} & \textbf{0.327} & 0.519 \\
\midrule
\multirow{3}{*}{\rotatebox[origin=c]{90}{\small ViT-S}}
 & DINOv3 + LoRA $r{=}32$        & $512{\times}512$ & 0.363 & 0.646 & 0.324 & 0.622 & 0.346 & 0.460 \\
 & DINOv3 + LoRA $r{=}64$        & $512{\times}512$ & 0.403 & 0.721 & 0.343 & 0.672 & 0.373 & 0.502 \\
 & DINOv3 + Full Fine-tuning      & $512{\times}512$ & 0.154 & 0.364 & 0.189 & 0.209 & 0.186 & 0.221 \\
\midrule
\multirow{3}{*}{\rotatebox[origin=c]{90}{\small ViT-B}}
 & DINOv3 + LoRA $r{=}32$        & $512{\times}512$ & 0.211 & 0.443 & 0.256 & 0.543 & 0.296 & 0.350 \\
 & DINOv3 + LoRA $r{=}64$        & $512{\times}512$ & 0.545 & 0.820 & 0.465 & 0.761 & 0.475 & 0.613 \\
 & DINOv3 + Full Fine-tuning      & $512{\times}512$ & 0.163 & 0.136 & 0.093 & 0.078 & 0.082 & 0.110 \\
\midrule
\multirow{3}{*}{\rotatebox[origin=c]{90}{\small ViT-L}}
 & DINOv3 + LoRA $r{=}32$        & $512{\times}512$ & \best{0.822} & \best{0.943} & \sbest{0.589} & \sbest{0.867} & \sbest{0.628} & \sbest{0.770} \\
 & DINOv3 + LoRA $r{=}64$        & $512{\times}512$ & \sbest{0.820} & \sbest{0.915} & \best{0.621} & \best{0.873} & \best{0.641} & \best{0.774} \\
 & DINOv3 + Full Fine-tuning      & $512{\times}512$ & 0.679 & 0.842 & 0.532 & 0.852 & 0.499 & 0.681 \\
\bottomrule
\end{tabular}
\end{table*}

\subsection{Robustness Evaluation}
\label{subsec:robustness}

\cref{tab:robust_noise,tab:robust_blur,tab:robust_jpeg} compare all DINOv3 configurations against prior methods on CASIAv1 under three post-processing perturbations: Gaussian noise (variance $\{3\text{--}23\}$), Gaussian blur (kernel $\{3\text{--}23\}$), and JPEG compression (quality $\{100\text{--}50\}$).
DINOv3 models achieve the highest average F1 under every perturbation type, with LoRA ViT-L consistently leading.
Across all methods, Gaussian blur induces by far the steepest degradation, while noise causes minimal drop—a pattern that reflects the spatial structure of the underlying representations and is analyzed in \cref{sec:discussion}.

\begin{table*}[t]
\centering
\caption{Robustness under Gaussian noise on CASIAv1. Pixel-level F1 at six noise variance levels. Prior method results from~\cite{mesorch2025}. Best per column highlighted in green.}
\label{tab:robust_noise}
\setlength{\tabcolsep}{5pt}
\begin{tabular}{@{}llcccccccc@{}}
\toprule
 & Method & Clean & 3 & 7 & 11 & 15 & 19 & 23 & Avg. \\
\midrule
\multirow{5}{*}{\rotatebox[origin=c]{90}{\small Prior work}}
 & MVSS-Net~\cite{Dong_2023}      & 0.5832 & 0.5824 & 0.5822 & 0.5764 & 0.5736 & 0.5620 & 0.5613 & 0.5744 \\
 & PSCC-Net~\cite{liu2022psccnet} & 0.6304 & 0.6127 & 0.5752 & 0.5540 & 0.5402 & 0.5232 & 0.5115 & 0.5639 \\
 & CAT-Net~\cite{kwon2022catnet}  & 0.8081 & 0.7979 & 0.7883 & 0.7832 & 0.7720 & 0.7573 & 0.7551 & 0.7802 \\
 & TruFor~\cite{guillaro2023truforleveragingallroundclues} & 0.8208 & 0.7666 & 0.7378 & 0.7190 & 0.6947 & 0.6832 & 0.6780 & 0.7286 \\
 & Mesorch~\cite{mesorch2025}     & 0.8398 & 0.8205 & 0.8050 & 0.7968 & 0.7887 & 0.7780 & 0.7696 & 0.7998 \\
\midrule
\multirow{3}{*}{\rotatebox[origin=c]{90}{\small ViT-S}}
 & DINOv3 + LoRA $r{=}32$  & 0.7870 & 0.7837 & 0.7716 & 0.7698 & 0.7584 & 0.7561 & 0.7517 & 0.7683 \\
 & DINOv3 + LoRA $r{=}64$  & 0.8025 & 0.7983 & 0.7901 & 0.7865 & 0.7752 & 0.7747 & 0.7692 & 0.7852 \\
 & DINOv3 + Full Fine-tuning        & 0.7043 & 0.6997 & 0.6946 & 0.6918 & 0.6891 & 0.6822 & 0.6789 & 0.6915 \\
\midrule
\multirow{3}{*}{\rotatebox[origin=c]{90}{\small ViT-B}}
 & DINOv3 + LoRA $r{=}32$  & 0.8395 & 0.8297 & 0.8148 & 0.8061 & 0.8007 & 0.7962 & 0.7871 & 0.8106 \\
 & DINOv3 + LoRA $r{=}64$  & 0.8630 & 0.8493 & 0.8379 & 0.8260 & 0.8204 & 0.8137 & 0.8150 & 0.8322 \\
 & DINOv3 + Full Fine-tuning        & 0.8148 & 0.8067 & 0.7945 & 0.7834 & 0.7787 & 0.7748 & 0.7719 & 0.7893 \\
\midrule
\multirow{3}{*}{\rotatebox[origin=c]{90}{\small ViT-L}}
 & DINOv3 + LoRA $r{=}32$  & 0.9071 & \rbest{0.9027} & \rbest{0.9009} & \rbest{0.8987} & \rbest{0.8967} & \rbest{0.8889} & \rbest{0.8921} & \rbest{0.8982} \\
 & DINOv3 + LoRA $r{=}64$  & \rbest{0.9077} & 0.8972 & 0.8957 & 0.8919 & 0.8902 & 0.8875 & 0.8868 & 0.8939 \\
 & DINOv3 + Full Fine-tuning        & 0.8816 & 0.8784 & 0.8760 & 0.8726 & 0.8698 & 0.8697 & 0.8678 & 0.8737 \\
\bottomrule
\end{tabular}
\end{table*}

\begin{table*}[t]
\centering
\caption{Robustness under Gaussian blur on CASIAv1. Pixel-level F1 at six kernel sizes. Prior method results from~\cite{mesorch2025}. Best per column highlighted in green.}
\label{tab:robust_blur}
\setlength{\tabcolsep}{5pt}
\begin{tabular}{@{}llcccccccc@{}}
\toprule
 & Method & Clean & 3 & 7 & 11 & 15 & 19 & 23 & Avg. \\
\midrule
\multirow{5}{*}{\rotatebox[origin=c]{90}{\small Prior work}}
 & MVSS-Net~\cite{Dong_2023}      & 0.5832 & 0.4587 & 0.3097 & 0.2369 & 0.1890 & 0.1571 & 0.1392 & 0.2962 \\
 & PSCC-Net~\cite{liu2022psccnet} & 0.6304 & 0.5410 & 0.4531 & 0.3156 & 0.1655 & 0.1140 & 0.0775 & 0.3282 \\
 & CAT-Net~\cite{kwon2022catnet}  & 0.8081 & 0.7512 & 0.6532 & 0.5435 & 0.4337 & 0.3142 & 0.2142 & 0.5312 \\
 & TruFor~\cite{guillaro2023truforleveragingallroundclues} & 0.8208 & 0.7508 & 0.6881 & 0.6032 & 0.4563 & 0.2741 & 0.1304 & 0.5320 \\
 & Mesorch~\cite{mesorch2025}     & 0.8398 & 0.7898 & 0.7081 & 0.6277 & 0.5328 & 0.4193 & 0.2940 & 0.6016 \\
\midrule
\multirow{3}{*}{\rotatebox[origin=c]{90}{\small ViT-S}}
 & DINOv3 + LoRA $r{=}32$  & 0.7870 & 0.7396 & 0.6694 & 0.5818 & 0.4529 & 0.3339 & 0.2525 & 0.5453 \\
 & DINOv3 + LoRA $r{=}64$  & 0.8025 & 0.7614 & 0.6974 & 0.6116 & 0.4860 & 0.3700 & 0.2941 & 0.5747 \\
 & DINOv3 + Full Fine-tuning        & 0.7043 & 0.6453 & 0.5584 & 0.4427 & 0.3136 & 0.2338 & 0.1618 & 0.4371 \\
\midrule
\multirow{3}{*}{\rotatebox[origin=c]{90}{\small ViT-B}}
 & DINOv3 + LoRA $r{=}32$  & 0.8395 & 0.8061 & 0.7650 & 0.6840 & 0.5455 & 0.3890 & 0.2697 & 0.6141 \\
 & DINOv3 + LoRA $r{=}64$  & 0.8630 & 0.8357 & 0.7866 & 0.7116 & 0.5285 & 0.3091 & 0.1851 & 0.6028 \\
 & DINOv3 + Full Fine-tuning        & 0.8148 & 0.7655 & 0.7112 & 0.6222 & 0.4947 & 0.3650 & 0.2632 & 0.5767 \\
\midrule
\multirow{3}{*}{\rotatebox[origin=c]{90}{\small ViT-L}}
 & DINOv3 + LoRA $r{=}32$  & 0.9071 & \rbest{0.8857} & \rbest{0.8476} & \rbest{0.8069} & \rbest{0.7218} & \rbest{0.5900} & \rbest{0.4786} & \rbest{0.7482} \\
 & DINOv3 + LoRA $r{=}64$  & \rbest{0.9077} & 0.8798 & 0.8371 & 0.7941 & 0.6670 & 0.4810 & 0.3556 & 0.7032 \\
 & DINOv3 + Full Fine-tuning        & 0.8816 & 0.8545 & 0.8176 & 0.7543 & 0.6284 & 0.4518 & 0.3224 & 0.6729 \\
\bottomrule
\end{tabular}
\end{table*}

\begin{table*}[t]
\centering
\caption{Robustness under JPEG compression on CASIAv1. Pixel-level F1 at six quality factors. Prior method results from~\cite{mesorch2025}. Best per column highlighted in green.}
\label{tab:robust_jpeg}
\setlength{\tabcolsep}{5pt}
\begin{tabular}{@{}llcccccccc@{}}
\toprule
 & Method & Clean & 100 & 90 & 80 & 70 & 60 & 50 & Avg. \\
\midrule
\multirow{5}{*}{\rotatebox[origin=c]{90}{\small Prior work}}
 & MVSS-Net~\cite{Dong_2023}      & 0.5832 & 0.5695 & 0.5446 & 0.5170 & 0.4906 & 0.4489 & 0.3888 & 0.5061 \\
 & PSCC-Net~\cite{liu2022psccnet} & 0.6304 & 0.6220 & 0.5789 & 0.4930 & 0.4518 & 0.3846 & 0.2869 & 0.4925 \\
 & CAT-Net~\cite{kwon2022catnet}  & 0.8081 & 0.7896 & 0.7858 & 0.7431 & 0.7230 & 0.6838 & 0.6132 & 0.7352 \\
 & TruFor~\cite{guillaro2023truforleveragingallroundclues} & 0.8208 & 0.8060 & 0.7938 & 0.7017 & 0.6852 & 0.6329 & 0.4942 & 0.7049 \\
 & Mesorch~\cite{mesorch2025}     & 0.8398 & 0.8312 & 0.8194 & 0.7716 & 0.7706 & 0.7285 & 0.6552 & 0.7738 \\
\midrule
\multirow{3}{*}{\rotatebox[origin=c]{90}{\small ViT-S}}
 & DINOv3 + LoRA $r{=}32$  & 0.7870 & 0.7772 & 0.7645 & 0.7211 & 0.7077 & 0.6693 & 0.5817 & 0.7155 \\
 & DINOv3 + LoRA $r{=}64$  & 0.8025 & 0.7954 & 0.7818 & 0.7464 & 0.7286 & 0.6861 & 0.5981 & 0.7341 \\
 & DINOv3 + Full Fine-tuning        & 0.7043 & 0.6924 & 0.6777 & 0.6287 & 0.6033 & 0.5598 & 0.4855 & 0.6217 \\
\midrule
\multirow{3}{*}{\rotatebox[origin=c]{90}{\small ViT-B}}
 & DINOv3 + LoRA $r{=}32$  & 0.8395 & 0.8316 & 0.8186 & 0.7846 & 0.7659 & 0.7355 & 0.6577 & 0.7762 \\
 & DINOv3 + LoRA $r{=}64$  & 0.8630 & 0.8548 & 0.8463 & 0.8117 & 0.7982 & 0.7843 & 0.7103 & 0.8098 \\
 & DINOv3 + Full Fine-tuning        & 0.8148 & 0.8059 & 0.7943 & 0.7570 & 0.7429 & 0.7106 & 0.6292 & 0.7507 \\
\midrule
\multirow{3}{*}{\rotatebox[origin=c]{90}{\small ViT-L}}
 & DINOv3 + LoRA $r{=}32$  & 0.9071 & \rbest{0.9058} & 0.8946 & 0.8755 & 0.8753 & 0.8598 & \rbest{0.8060} & 0.8749 \\
 & DINOv3 + LoRA $r{=}64$  & \rbest{0.9077} & 0.9035 & \rbest{0.8959} & \rbest{0.8775} & \rbest{0.8758} & \rbest{0.8622} & 0.8046 & \rbest{0.8753} \\
 & DINOv3 + Full Fine-tuning        & 0.8816 & 0.8778 & 0.8749 & 0.8485 & 0.8459 & 0.8312 & 0.7716 & 0.8474 \\
\bottomrule
\end{tabular}
\end{table*}

\section{Discussion}
\label{sec:discussion}

\paragraph{Parameter efficiency of LoRA.}
Using 11--34$\times$ fewer trainable parameters (\cref{subsec:adaptation}), LoRA consistently outperforms full fine-tuning across all backbone scales; we attribute this to preserving general-purpose representations learned during DINOv3 pre-training: constraining backbone updates to a low-rank subspace retains the broad feature diversity needed for cross-dataset generalization, whereas full fine-tuning risks overfitting to the specific manipulation artifacts in the training data.
The robustness results (\cref{tab:robust_noise,tab:robust_blur,tab:robust_jpeg}) support this interpretation: LoRA models degrade more gracefully under post-processing perturbations, indicating their representations are not narrowly tuned to training-distribution statistics.

\paragraph{Backbone scaling.}
Performance scales monotonically from ViT-S to ViT-L under both adaptation strategies, with NIST16 improving by 17.4 absolute F1 points from ViT-S to ViT-L LoRA $r{=}32$.
Larger ViT backbones encode richer spatial and semantic structure within each $16{\times}16$ patch token, which may help the decoder resolve finer manipulation boundaries and more subtle forensic traces.
The lower-rank configuration ($r{=}32$) outperforms $r{=}64$ for ViT-L (0.847 vs.\ 0.837), suggesting larger models already encode sufficient task-relevant information in their frozen features, requiring less low-rank adaptation.
This trend holds across training protocols: under the MVSS-Net setup, LoRA $r{=}64$ scales consistently from ViT-S (0.502) to ViT-B (0.613) to ViT-L (0.774), suggesting that backbone capacity transfers regardless of training-data volume.

\paragraph{Dataset-specific observations.}
Columbia's uncompressed format and clean binary masks make it the easiest target, with all models achieving F1 $\geq$ 0.887.
NIST16 contains native JPEG compression that creates grid-aligned artifacts resembling manipulation boundaries, confounding all detectors and capping F1 at 0.636.
Coverage tests the ability to distinguish subtle copy-move duplications; the steep scaling curve (+25.9 F1 points from ViT-S to ViT-L) indicates this task benefits disproportionately from richer feature representations.
CASIAv1, whose training counterpart (CASIA2) is in the CAT-Net training mix, shows that LoRA's advantage persists even at close domain proximity.
IMD2020, a genuine out-of-domain test under the MVSS-Net protocol, is the most demanding: prior methods peak at 0.327 (IML-ViT), while LoRA ViT-L reaches 0.641, nearly doubling the previous best and providing the strongest evidence for cross-domain generalization.

\paragraph{Full fine-tuning collapse under data-scarce training.}
Under the MVSS-Net protocol, adaptation strategy becomes decisive.
Under large-scale multi-source supervision, full fine-tuning is competitive, trailing LoRA ViT-L by only 2.1 F1 points (0.826 vs.\ 0.847; \cref{tab:main_results}).
Restricted to 5{,}123 training images, however, full fine-tuning collapses: ViT-S and ViT-B train unstably and peak early, yielding poor average F1 of only 0.221 and 0.110.
ViT-L partially recovers (0.681) but lags its LoRA counterpart by 9.3 points (\cref{tab:mvss_results}).
This regime-dependent behavior is consistent with the mechanism described above: scarce supervision exposes full fine-tuning's high-dimensional parameter space, which may cause it to overwrite the pre-trained priors that LoRA's low-rank constraint preserves.
Even so, FullFT ViT-L (0.681) still exceeds the strongest prior method (TruFor, 0.530), underscoring that DINOv3 representations carry substantial forensic information even under suboptimal adaptation.

\paragraph{Robustness patterns and DINOv3 representations.}
Gaussian noise causes minimal degradation: for LoRA ViT-L $r{=}32$, F1 drops only 1.7\% (from 0.907 to 0.892) at the highest severity, suggesting that DINOv3 features encode information above pixel-level statistics.
JPEG re-compression produces moderate degradation (11.1\% relative drop to 0.806 at quality 50), likely because it partially disrupts the compression-grid discontinuities at manipulation boundaries.
Gaussian blur is devastating, reducing F1 by 47.2\% (from 0.907 to 0.479 at kernel size 23), possibly because it smooths boundary cues across adjacent $16{\times}16$ patches.
This asymmetry is consistent with ViT's patch-based tokenization: within-patch frequency information is aggregated into each token, which may confer natural noise tolerance, but blur spreads information across patch boundaries, potentially corrupting the spatial discontinuities that manipulation detectors rely upon.

\section{Conclusion}
\label{sec:conclusion}

In this report, we presented a simple but effective DINOv3-based baseline for image manipulation detection and localization.
With LoRA adaptation and only a lightweight convolutional decoder, our approach substantially outperforms existing specialized IMDL methods on standard benchmarks, improving the average pixel-level F1 score by up to 17.0 points under the CAT-Net protocol.
Under the more constrained MVSS-Net setting, DINOv3 with LoRA also remains the top-performing configuration across all five benchmarks.

We consistently find that LoRA is more effective than full fine-tuning, especially in low-data regimes.
When trained only on CASIAv2, full fine-tuning becomes highly unstable, while LoRA continues to transfer well across datasets.
These results suggest that, for IMDL, preserving strong pre-trained representations is crucial for robust adaptation and generalization.

Looking forward, we believe progress in IMDL will benefit not only from stronger foundation-model-based baselines, but also from larger, more diverse, and more data-centric datasets.
We hope this work can provide a useful reference point for future research and a practical starting point for real-world image-forensic applications.
{
    \small
    \bibliographystyle{ieeenat_fullname}
    \bibliography{main}
}


\end{document}